# Explainable Artificial Intelligence to Detect Image Spam Using Convolutional Neural Network

Zhibo Zhang
C2PS, Department of Electrical
Engineering and Computer Science
Khalifa University
Abu Dhabi, United Arab Emirates
qiuyuezhibo@gmail.com

Ernesto Damiani
C2PS, Department of Electrical
Engineering and Computer Science
Khalifa University
Abu Dhabi, United Arab Emirates
ernesto.damiani@ku.ac.ae

Hussam Al Hamadi
C2PS, Department of Electrical
Engineering and Computer Science
Khalifa University
Abu Dhabi, United Arab Emirates
hussam.alhamadi@ku.ac.ae

Chan Yeob Yeun
C2PS, Department of Electrical
Engineering and Computer Science
Khalifa University
Abu Dhabi, United Arab Emirates
chan.yeun@ku.ac.ae

Fatma Taher
Department of Computing and Applied Technology
Zayed University
Dubai, United Arab Emirates
Fatma.Taher@zu.ac.ae

*Abstract*—Image spam threat detection has continually been a popular area of research with the internet's phenomenal expansion. This research presents an explainable framework for detecting spam images using Convolutional Neural Network (CNN) algorithms and Explainable Artificial Intelligence (XAI) algorithms. In this work, we use CNN model to classify image spam respectively whereas the post-hoc XAI methods including Local Interpretable Model Agnostic Explanation (LIME) and Shapley Additive Explanations (SHAP) were deployed to provide explanations for the decisions that the black-box CNN models made about spam image detection. We train and then evaluate the performance of the proposed approach on a 6636 image dataset including spam images and normal images collected from three different publicly available email corpora. The experimental results show that the proposed framework achieved satisfactory detection results in terms of different performance metrics whereas the model-independent XAI algorithms could provide explanations for the decisions of different models which could be utilized for comparison for the future study.

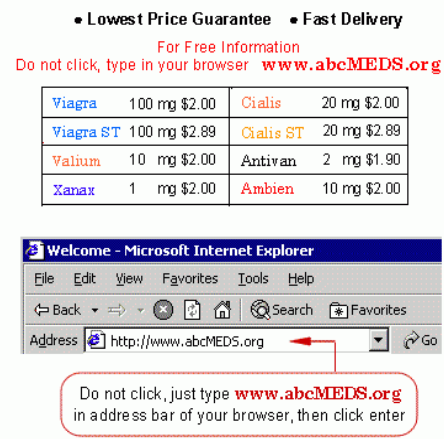

Fig. 1. Sample image spam

## I. INTRODUCTION

Email is now the formal communication type that is used the most frequently by the vast majority of internet users. Spam email, has, however, grown to be a significant issue for cyber security in recent years due to the increased use of email in routine commercial transactions and general communication [1]. While over 306.4 billion emails were sent and received per day in 2021, spam email communications made up nearly half of all email traffic [2]. Although there are several mechanisms passively filtering these spam emails in the mailbox to stop them, attackers continue to use numerous strategies to evade these anti-spam systems. For instance, image spam has been widely utilized by attackers to evade the detection of text-based spam filtering systems. Furthermore, compared to text-based spam, image spam contains more complicated suspicious information which is more damageable for users [3].

Although Machine Learning (ML) based spam detectors deploying ML algorithms including Support Vector Machines (SVM), Naïve Bayes (NB), Random Forest (RF), etc achieved great performance for filtering text-based email spam [4], image spam could show text content primarily as an image shown in Fig. 1. Besides the modifying techniques such as utilizing various colors, altering letter size, and introducing speckles, spam images may contain illustrations in addition to text as well.

Conventionally, Optical Character Recognition (OCR) [5] was utilized by anti-spam systems to extract and recognize the words embedded in the image spam and transform these images into texts. Traditional text-based spam detectors could be deployed to detect the image spam processed by the OCR mechanisms. Nevertheless, most OCR techniques are prone to errors whereas the image's quality has a big impact on how well the OCR techniques perform [6]. On the other hand, spam attackers improved their strategies, such as changing the background and foreground colors, text font styles, and image rotation, to circumvent the OCR-based image spam detection mechanisms.

In [7], the authors utilized OCR-based systems to recognize the inserted Chinese text in the image spam, then a text-based classifier could be used to distinguish between spam and ham emails. Moreover, to address the inadequacies of the OCR techniques, a keyword reconstruction approach based on Word Activation Force (WAF) model is presented in this paper. And the experimental outcomes on an individual dataset of image spam (which is publicly accessible) confirm the effectiveness of the proposed technique, which outperformed the original OCR systems in real-world usage with a complicated background in image spam.

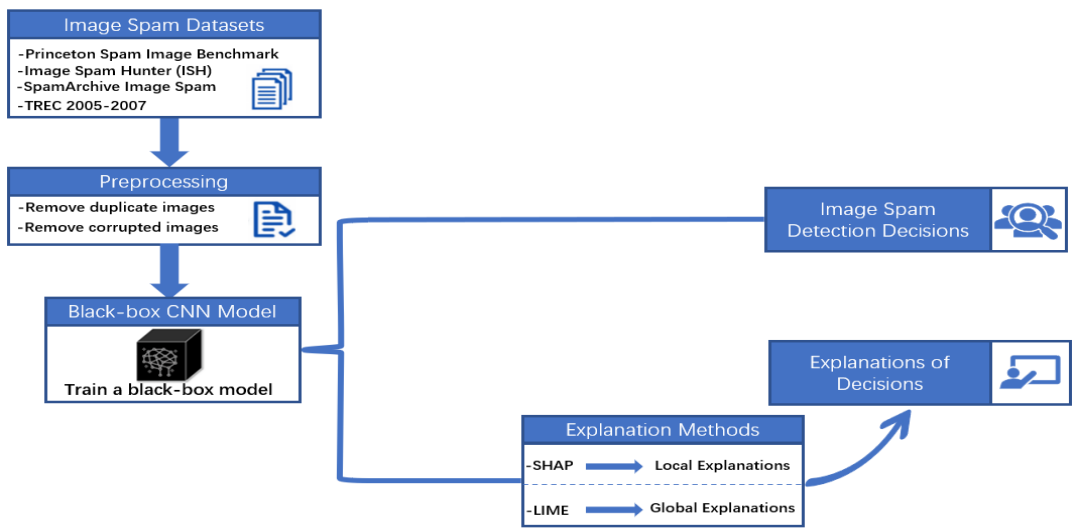

Fig. 2. Proposed XAI-based spam image detector framework

On the other hand, to counter the drawback of the OCR-based Machine Learning text spam detectors, Deep Learning techniques including CNN were introduced to address the issue of image spam detection as well. In [8], Sriram et al. utilized 3 different datasets to train 2 CNN models as well as a few pre-trained ImageNet architectures including VGG19 and Xception. To address data imbalance, the impact of using a cost-sensitive learning strategy was investigated in this study as well. In the best situation, some of the models that are suggested achieved accuracy levels of up to 99%. In [9], a dataset with 810 normal images and 928 spam images was utilized to build a CNN-based spam detection system and an accuracy of over 90% was achieved. In this work, the manual feature extraction task was avoided so that the time and effort for processing the images was decreased compared with other conventional image processing techniques.

However, while there have been studies using deep learning approaches to detect image spam, very few research focus on the trustworthiness and explainability of deep learning-based image spam detectors. On the other hand, due to the black-box nature of deep learning algorithms such as CNN, XAI has been widely utilized in other domains of image classification tasks including explainable medical image classification [10], face recognition[11], and autonomous driving system [12]. To increase the transparency of the deep learning approaches and provide the user confidence about the system decision, especially in sensitive areas such as medical diagnosis and cyber security, different methods have been deployed to visualize, interpret and explain deep learning models.

Ran et al. [10] presented a comprehensive attention-based CNN (CA-Net) architecture for more precise and explainable medical image segmentation. Compared with the conventional CNN-based automatic medical image segmentation approaches, the proposed CA-Net framework enhanced the model explainability by visualizing the attention weight maps.

Other than that, XAI approaches were utilized in other cyber security tasks such as intrusion detection as well. Zakaria et al. [13] designed a novel Deep Learning and XAI-based system for intrusion detection in IoT networks. Three different explanation methods including LIME, SHAP, and RuleFit were deployed to provide local and global explanations for the single output of the DNN model and the most significant features conducted to the intrusion detection decision respectively.

Therefore, to bridge this gap of introducing explainability to image spam detection, the main contributions proposed in this paper are listed as follows:

(1) Proposing an explainable image spam detection framework using CNN models and XAI algorithms LIME and SHAP.

(2) Introducing explainability for the decisions made by the black-box CNN models to make the spam detection process more transparent to users.

(3) Comparing and investigating the explainable results of image spam samples that were not detected accurately to provide information for future studies.

The rest of this paper is organized as follows: Section II introduces the proposed explainable framework for the detection of image spam using CNN models. Section III provides experimentation results and analysis in terms of conventional performance metrics as well as explainability. Section IV concludes this paper and provides prospects for future work.

II. METHODOLOGY

In this section, the methods including the pre-processing, CNN model, Local Interpretable Model-Agnostic Explanations (LIME), and Shapley Additive Explanations (SHAP) are introduced to build the XAI-empowered image spam detection framework. An overview of the proposed architecture is shown in Fig. 2 and the different stages of the XAI-based image spam detector are described in the subsections respectively.

A. Pre-processing

The first step of the proposed method is the pre-processing of the spam images. In this work, we utilized three spam image datasets and split the datasets into the training set and testing set. However, there are numerous duplicate images and corrupted images in the datasets deployed. Other than that, there is no standard format for the images in the dataset. Some images are JPG types whereas some images are PNG or GIF types. Therefore, we omitted the duplicate and corrupted images first and then transformed all images into the JPG types.

*B. The CNN model*

In this study, a CNN model is designed for the classification of the image parts of spam e-mails. For the hyperparameters used in the proposed CNN model, the learning rate is set to be 0.0001, the optimizer algorithm is chosen as RMSprop, the epoch is set to be 30, and the batch size is 20.

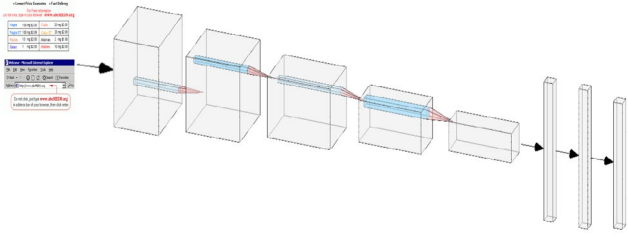

Fig. 3. CNN model utilized in this work

As shown in Fig. 3, before being fed into the CNN model, the image files are decoded from JPG content to RGB grids of pixels 256 × 256 and converted into floating-point tensors after that. Then, the pixel values between 0 and 255 are rescaled to [0,1], and the input shape is transformed into (128 × 128 × 3). The derived CNN model has 4 convolution layers of filter sizes 32, 64, 128, and 128 respectively. And every convolutional layer is followed by a Max Pooling layer with pooling size 2 and the activation function ReLU. Finally, using a sigmoid activation function, a dense layer of a single neuron is utilized. The detailed structure of the CNN model is represented with layer details in Table I.

TABLE I. DETAILED ARCHITECTURE OF CNN MODEL

| Layer Type | Output Shape | Parameter Number |
|---|---|---|
| Conv2D(32, (3,3)) | (None, 126, 126, 32) | 896 |
| MaxPooling2D ((2,2)) | (None, 63, 63, 32) | 0 |
| Conv2D(64, (3,3)) | (None, 61, 61, 64) | 18496 |
| MaxPooling2D ((2,2)) | (None, 30, 30, 64) | 0 |
| Conv2D(128, (3,3)) | (None, 28, 28, 128) | 73856 |
| MaxPooling2D ((2,2)) | (None, 14, 14, 128) | 0 |
| Conv2D(128, (5,5)) | (None, 10, 10, 128) | 409728 |
| MaxPooling2D ((2,2)) | (None, 5, 5, 128) | 0 |
| Flatten | (None, 3200, 1, 1) | 0 |
| Dense(512) | (None, 512) | 1638912 |
| Dense(1) | (None, 1) | 512 |

*C. LIME explanation model*

LIME stands for Local Interpretable Model-agnostic Explanations proposed by Marco et al. in [14], Finding an interpretable model over the interpretable representation that is locally faithful to the classifier and intelligible to humans is the core objective of the LIME algorithm.

Define the explanation model as $g \in G$, where G is a group of interpretable models that can be displayed to a user graphically (e.g., linear model). $\pi_x(z)$ is then utilized to represent the proximity between the instance z and x to define the locality around x. Then define an objective function ξ(x), and the L-function in ξ(x) serves as a metric describing how the infidelity $g$ approximates f (complex model) through $\pi_x(z)$ in the local definition. And the L-function is minimized to obtain the optimal solution of the objective function when Ω(g) (the explanatory model complexity) is low enough to be understood by humans. The explanation function ξ(x) produced by the LIME algorithm is as follows:

$$\xi(x) = \arg\min_{g \in G} L(f, g, \pi_x(z)) + \Omega(g) \quad (1)$$

The formula for calculating the similarity degree $\pi_x(z)$ is as follows

$$\pi_x(z) = \exp\left(-\frac{D(x,z)^2}{\sigma^2}\right) \quad (2)$$

With the definition of the similarity degree $\pi_x(z)$ in equation 2, the original objective function can be rewritten in the following form in equation 3. Where f(z) is the predicted value of the perturbed sample, on the d-dimensional space (original features), and takes that predicted value as the answer, and g(z′) is the predicted value on the d'-dimensional space (interpretable features), and then uses the similarity as the weight, so the above objective function can then be optimized by linear regression.

$$\xi(x) = \sum_{z,z' \in Z} \pi_x(z) \, (f(z) - g(z'))^2 \quad (3)$$

*D. SHAP explanation model*

Known as a unifying framework for the interpretation of the black-box models, SHAP, which stands for Shapley Additive exPlanations, was introduced by Scott et al. in [15]. The goal of SHAP is to explain the prediction of instance x by calculating the contribution of each feature to the prediction x. The SHAP explanation method calculates the Shapley value based on coalition game theory. The feature values of data instances act as participants in the coalition (set). The Shapley values tell us how to fairly distribute the "expenditures" (i.e., predictions) among the features. A player can be a single feature value, for example, for tabular data. A player can also be a set of feature values. SHAP specifies the interpretation as the following equation:

$$g(z') = \phi_0 + \sum_{j=1}^{M} \phi_j z'_j \quad (4)$$

Where $g$ stands for the explanation model, $z' \in \{0,1\}^M$ stands for the coalition vector, $M$ stands for the maximized coalition size, and $\phi_j \in R$ stands for the feature attribution of feature j.

## III. EXPERIMENT RESULTS AND ANALYSIS

This paper simulated the performance of the models proposed in Section II in the environment of Python 3.8. The experiment is carried out in the operating system of Windows 10, 4 cores CPU, 8.00 GB RAM, and 4G GPU.

*A. Data Set*

To train and evaluate the performance of the proposed XAI-based CNN image spam email detection models, this paper implements four publicly available image email datasets for the experiments, including image spam and normal images respectively. Many anonymous individuals contributed to the construction of the utilized "SpamArchive spam" dataset [16]. The image spam hunter (ISH) dataset was generated by a team at Northwestern University [17], which includes jpg-formatted actual images that are spam. Images of spam were gathered for the Princeton Spam Image Benchmark [18] from various email accounts. For these datasets, after the pre-processing stages of removing the corrupted and duplicate images, we were left with a dataset of 6636 samples for experiments. Below Table II shows the specifics of the datasets utilized in the experiments.

TABLE II. DATASETS USED IN SIMULATION

| Type | Datasets | Number |
|---|---|---|
| Spam | Princeton Spam Image Benchmark | 788 |
| | Image Spam Hunter (ISH) | 1063 |
| | SpamArchive Image Spam | 1354 |
| | TREC 2005-2007 | 759 |
| | Total | 3964 |
| Normal | Princeton Spam Image Benchmark | 554 |
| | Image Spam Hunter (ISH) | 759 |
| | SpamArchive Image Spam | 810 |
| | TREC 2005-2007 | 549 |
| | Total | 2672 |

### B. Statistical Metrics

In this section, in terms of evaluating the performance of the detection models, the confusion matrix is used where FP, FN, TP, and TN are defined as follows. False Positive (FP) is the number of misclassified legitimate emails, False Negative (FN) is the amount of misclassified spam, True Positive (TP) is the number of properly classified spam, and True Negative (TN) is the number of correctly classified legitimate emails. Based on these, the statistical metrics including accuracy, recall, precision, and F1-score are defined as the following equations:

$$Accuracy = \frac{TP + TN}{TP + TN + FP + FN} \quad (5)$$

$$Recall = \frac{TP}{TP + FN} \quad (6)$$

$$Precision = \frac{TP}{TP + FP} \quad (7)$$

$$F1 - Score = \frac{2 \times (Precison \times Recall)}{Precison + Recall} \quad (8)$$

### C. Results and Discussion

In this section, we present our assessment results based on the statistical metrics discussed above. In addition, we provide some analysis and discussion of the experimental findings.

Fig. 4. Confusion matrix of the proposed detector

As we discussed in the last section, there are 6636 samples in the dataset, and we divided the dataset into the training set and testing set with a ratio of 3:1. Therefore, as shown in Fig.4, the confusion matrix of the proposed XAI-based CNN spam image detector is presented. From the confusion matrix, the statistical metrics including accuracy, recall, precision, and F1-score could be calculated as 97.16%, 95.68%, 98.79%, and 97.16%, which are acceptable compared to other benchmark algorithms shown in Table III.

TABLE III. THE PERFORMANCE METRICS OF THE MODELS

| Datasets | Accuracy | Recall | Precision | F1-Score |
|---|---|---|---|---|
| Proposed Framework | 97.16% | 95.68% | 98.79% | 97.16% |
| [8] DCNN | 97.1% | 98,1% | 96.3% | 97.2% |
| [9] CNN | 91.7% | 85.7% | 100% | 92.3% |

However, the explainability of the testing samples could be expanded through heatmap, LIME, and SHAP approaches in the proposed XAI-CNN framework. As shown in Fig. 5, SHAP values provided an additive measure of feature importance. In the context of an image, each pixel is treated as a feature, therefore, SHAP values can be used to determine the pixel level importance in classifying images Fig. 5 showed the samples of 2 spam images and 1 normal image.

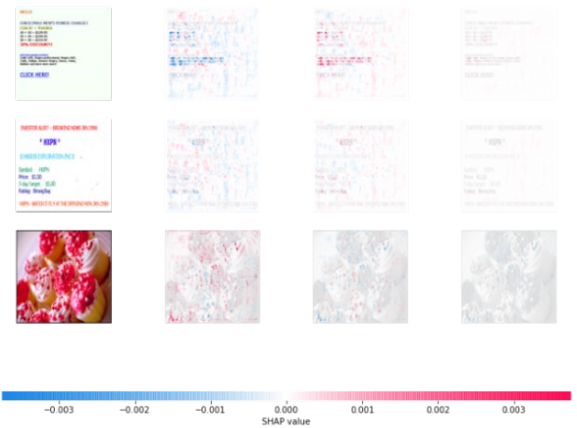

Fig. 5. SHAP explainability for pixel importance of sample images

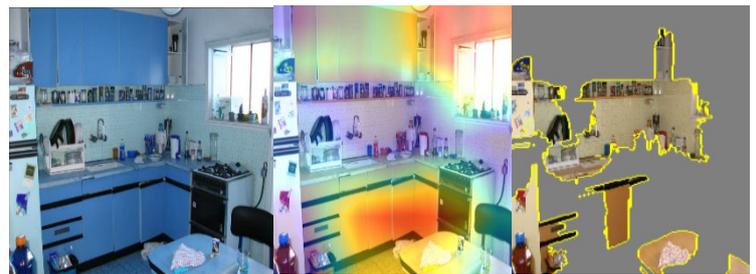

Fig. 6. Heatmap and LIME explainability of a normal image

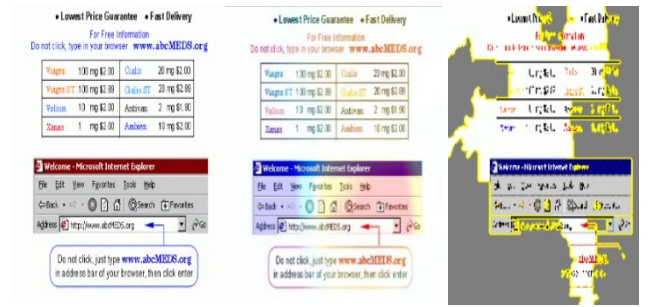

Fig. 7. Heatmap and LIME explainability of a spam image

On the other hand, Fig. 6 and Fig. 7 could show the explainability provided by the heatmap and the LIME methods experimented on a spam image sample and a normal image sample. LIME method works by making alterations to different features on a particular input and seeing which of those alterations make the biggest difference to the output classification. Thus highlighting the features most relevant to the network's decision. The key to LIME's effectiveness is local elements. That means that it does not try to explain all the decisions that a network might make across all possible inputs, only the factors that use to determine its classification for one particular input.

## IV. Conclusion and Future work

To avoid image spam evading the conventional text-based spam e-mail filters and provide a robotic and explainable filtering service in the cyber security area, this paper introduces an XAI-CNN model for an image spam email filtering system. After the decisions made by the CNN model, the proposed framework integrated two different methodologies of XAI (i.e., SHAP and LIME) to increase the explainability, transparency, and user trust in the CNN-model decisions. To demonstrate the efficiency of the XAI-CNN model, we utilized 4 image spam datasets to compose a large dataset for experiments. The experimental findings demonstrated the effectiveness of our proposed framework in detecting image spam as well as incorporating additional information and an explanation of how and why the CNN model makes such detection judgments.

In future work, we plan to explore the explainability of the OCR-transformed image spam in the context of a text-based detector and compare the differences between interpretability differences of the image-based and text-based frameworks. Furthermore, user interfaces will be developed to provide explainabilities and transparency of decisions based on multiple XAI methods.